\documentclass[review]{elsarticle}
\usepackage{lineno,hyperref}
\modulolinenumbers[5]
\usepackage{graphicx}
\usepackage{amsmath}
\usepackage{mathtools}
\usepackage{float}
\usepackage[utf8]{inputenc}
\usepackage[british]{babel}
\usepackage{tabularx}
\usepackage{multirow}
\usepackage{picins}
\journal{Expert Systems with Applications}
\biboptions{authoryear}
\begin{document}

\begin{frontmatter}

\title{A Convolutional Feature Map based Deep Network targeted towards Traffic Detection and Classification}


\author[a]{Baljit Kaur\corref{mycorrespondingauthor}}
\cortext[mycorrespondingauthor]{Corresponding author at: Department of Computer Science and Engineering, Thapar Institute of Engg and Tech, Patiala, Punjab, India}
\ead{baljitkaur13@gmail.com}
\author[a]{Jhilik Bhattacharya}
\ead{jhilik@thapar.edu}
\address[a]{Department of Computer Science and Engineering, Thapar Institute of Engg and Tech, Patiala, Punjab, India}

\begin{abstract}
This research mainly emphasizes on traffic detection thus essentially involving object detection and classification.  The particular work discussed here is motivated from unsatisfactory attempts of re-using well known pre-trained object detection networks for domain specific data. In this course, some trivial issues leading to prominent performance drop are identified and ways to resolve them are discussed. For example, some simple yet relevant tricks regarding data collection and sampling prove to be very beneficial. Also, introducing a blur net to deal with blurred real time data is another important factor promoting performance elevation. We further study the neural network design issues for beneficial object classification and involve shared, region-independent convolutional features. Adaptive learning rates to deal with saddle points are also investigated and an average covariance matrix based pre-conditioned approach is proposed. We also introduce the use of optical flow features to accommodate orientation information. Experimental results demonstrate that this results in a steady rise in the performance rate.
\end{abstract}

\end{frontmatter}


\section{Introduction}
Vehicle detection methods can be categorised as: Moving-vehicle detection on the basis of background estimation or cascade based object detection. In the former case vehicle candidates can be extracted from foreground blocks, obtained by removing the predicted background from original input images \citep{zhou2007moving} \cite{vargas2010enhanced}. Standard histogram-based contrast (HC) \cite{li2017vehicle} or Bayesian probability models \cite{yao2017coupled} are some examples of the same. These methods have low computational complexity and can be applied to simple and stable background based applications. However, they are not appropriate for coping with congested urban traffic scenes specially during slow traffic movement due to lack of flow information. Alternatively, the movement of the object cannot be determined without taking its inherent information into consideration. Cascade based detection techniques involve searching the whole image region wise to test for vehicle candidature via suitable feature extraction techniques. Due to availability of computer resources, and high performance rate of deep CNN, ConvNets are replacing hand crafted feature extraction. Among CNNs for object detection, R-CNN, fast RCNN, faster RCNN are widely adapted. R-CNN extract region proposals, compute CNN features and classify the objects. To improve  computation ability, Fast R-CNN used region of interest pooling by sharing the forward pass of CNN. These region proposals were created using selective search which was replaced by a RPN in faster R-CNN \citep{ren2015faster}. Here a single network composed of region proposal and Fast R-CNN was used by sharing their convolutional features. An option to add segmentation properties to Fast RCNN was enabled by putting an object mask predicting feature with the already occurring branch for bounding box recognition \citep{he2017mask}. It was noticed that these object detection networks utilized VGG16 with PASCAL dataset.  One reason behind the choice could be that it is very deep network with 41 layers. Hence, it is better from AlexNet in terms that it has large kernel-sized filters having 11 in the first and 5 in the second convolutional layer with a couple of 3X3 sized kernel filters one after the other. With a given effective area size of input image on which output depends, couple of smaller sized kernel is better than large sized kernel because more than one non-linear layers result the deep network which makes it possible to learn more complex features at a lower cost. GoogleNet on the contrary has quite a different architecture: it uses combinations of inception modules, each including some pooling, convolutions at different scales and concatenation operations. GoogleNet and ResNet do not allow region wise classifiers due to absence of fully connected layers. However, another widely used network YOLO \citep{redmon2016yolo9000} composed of entirely convolutional layers trained and tested on PASCAL VOC and COCO datasets proved to be quite accurate and fast.\\
In the present context, we engaged faster RCNN for object detection from self captured videos. Our goal was to detect traffic regions from a scene and further use this for a dedicated vehicle classification CNN. However we failed to obtain satisfactory detection results, leading us to explore the possible causes.(i) The primary reason could be a problem with the dataset we captured. For example size of the dataset may be too less to obtain suitable results. Also for using video data, a lot of similar frames were captured hence resulting in non homogeneity. This will result in low gradients during learning leading to slow or no convergence. (ii) Secondly, blurred images captured due to random movement of camera can have negative impact on the network. So we need to carefully eliminate blurred images or remove blur from images \citep{xu2013unnatural}, \citep{cho2012text}, \citep{zheng2013forward}. Also the network can be trained using blurred images only. An example of the same is \citep{guo2017vehicle} where GoogleNet was used to train blurred data to improve results. (iii) Third, the design of deep network for object classification plays an important role. While exploring the network design for classification, we studied that apart from deep feature maps, a deep and convolutional per-region classifier has special importance for object detection. It was argued by several researchers that models for image classification such as GoogLeNets and ResNets did not give good detection accuracy without the use of per-region classifier.(iv) Another important part could be the features provided to the network for classification. We know that for different object detection challenges, deep neural networks improve performance by averaging over different crops or scale of a particular image. PCA and whitening of pixels was used to reduce the overfitting problem in Imagenet. This caters for intensity variations in the training image. Some work considered use of RGB images with depth data for improving the accuracy of object detection  \citep{cao2017exploiting}\citep{niessner2017investigations}. These include training a network from scratch using RGB,depth and/or LIDAR data or finetuning pretrained nets like VGG/Alex-net with depth/LIDAR data for improving object detection performance. In the current work, we consider using orientation value obtained from Optical flow with the idea that it will encapsulate pose information. Optical flow has been used along with hand crafted HOG and LUV features for pedestrian detection on Caltech-USA pedestrian dataset\citep{rauf2016pedestrian}. Occlusion edge detections using optical flow has also been reported by researchers \citep{pop2017incremental}  \citep{sarkar2017deep}. They trained CNN with Intensity, Depth and Flow images for each frame. An approach based on Optical flow with the combination of deep learning for visual odometry has been proposed by Muller and Savakis \citep{muller2017flowdometry}. In this, Optical flow images are used as input to a CNN, which calculates a rotation and displacement for each image pixel. The displacements and rotations were applied incrementally to construct a map of where the camera has traveled. CNN trained with optical flow for vehicle detection and classification is yet to be found. In this regard we have the following \textbf{\emph{contributions}} towards developing a traffic detection application.
\begin{enumerate}
    \item \textbf{Suitable pre-processing of data-set to remove homogeneity.}
    \item \textbf{A Network on Convolutional Feature Maps(NoC) trained and used for classification. We perform extensive experiments with different learning rates and layer designs to understand how the learning is affected.}
    \item \textbf{Use of blur NoCs trained particularly with blurred dataset to accommodate blurred scenes during real-time processing.}
    \item \textbf{We use a multimodal fusion, where we fuse the convolution feature maps of individual columns of the multicolumn CNN using summation operation. This proves to be beneficial by accommodating multimodal features with minimal computational space and speed as opposed to fusion techniques via concatenation. For example in our case we fuse features extracted from 5th convolutional layer of pretrained network using RGB(Intensity) images and orientation features via optical flow images; represented as $\{conv5\_F_{I},conv5\_F_{O}\}$ respectively. }
    \item \textbf{We propose an average covariance based pre-conditioning approach to deal with saddle points in deep networks.}
\end{enumerate}
The rest of paper is framed as related literature for the proposed work is discussed in Section 2. Data collection and data preprocessing are elaborated in Section 3. Experiments done for the proposed work are presented in Section 4. Finally, Section 5 concludes the whole work.

\section{Related Work}
Considerable amount of work is reported on candidate region detection as well as classification on different categories of images. While some research is focused on application specific detection tasks as detailed in (a), some others mainly focus on improving detections with respect to speed, accuracy, false alarms detection as elaborated in (b).\\

(a) CNN based detection and classification techniques were implemented for detecting pedestrian, cyclist, vehicle type, animals and many more. A combined Framework for Concurrent detection of Pedestrian and Cyclist was proposed by \citep{li2017unified} using RCNN on upper body regions detected with ACF,LCDF \citep{nam2014local}. \citep{huo2016vehicle} classified different vehicle types(car, truck, bus and van) from different views using a multi-task RCNN. An animal detection technique using multilevel graph cut for combination motion with spatial context was presented in \citep{zhang2016animal}. The feature description for animal detection used was a combination of deep learning (pretrained caffe CNN) and oriented gradient histogram features encoded with Fisher vectors. \citep{zhuo2017vehicle} fine-tuned their own vehicle dataset using  GoogLeNet, pretrained with ILSVRC-2012 data, to obtain vehicle classification results. \citep{yao2017coupled} have detected vehicles using Bayesian probability model and classified multivehicle by adopting AlexNet pre-trained with ILSVRC 2012 ImageNet data set as classifier. Along with classification their framework also detected vehicle location. \citep{wang2016vehicle} detected vehicles using pre-trained fast-RCNN network and classified them into types via VGG\_CNN\_M\_1024 model. A K-means algorithm was utilized for clustering the vehicle data prior training of VGG\_CNN\_M\_1024. \\

(b) While most of the available research focus on using different networks and classifiers for specific applications, some researchers have particularly focused on increasing speed and accuracy while reducing false alarms for these. For example \citep{zhang2016accelerating} have presented an accelerating method that  proved to be effective for very deep models. They proposed a response reconstruction method that takes into account the nonlinear neurons and a low-rank constraint. A solution based on Generalized Singular Value Decomposition (GSVD) was developed for this nonlinear problem, without the need of SGD. Their method was evaluated under whole-model speedup ratios. It could effectively reduce the accumulated error of multiple layers due to the nonlinear asymmetric reconstruction. A method to reduce false alarms was introduced by \citep{kang2017t}  where detection results were  propagated to adjacent frames according to motion information. The resulted duplicate boxes were removed by non-maximum suppression (NMS). Another effective approach to reduce false alarms including Context based CNN object detection model was introduced by \citep{li2017attentive}. \citep{kang2017optimizing} have proposed a NOSCOPE  system for the purpose of accelerating neural network for video with the help of inference-optimized model search.


\section{Data Set}
The quality of data plays an important role in training any deep network. In general, the majority of reported research utilizes the bottleneck layer of a pre-trained network trained over millions of images for feature extraction purposes. For suiting this to the particular application, domain specific data is used for transfer learning or fine tuning. Collection of these comes with practical problems and needs to be dealt with, prior    to their usage. However most works directly discuss the data application and do not elaborate on the common prior problems and how they were tackled. Some potential problems faced while collecting data for classification are discussed so that this can be taken as a point of reference for future use of researchers. For the proposed system, outdoor traffic data including cars, pedestrians two wheelers etc is required in the form of videos. It is a well-known fact that data non-homogeneity in the form of various lighting, postures and other structural and environmental conditions is required for a robust classification. However experimental results have shown that poor quality data have a greater impact over false alarms as compared to the true accepts. Also maintaining non-homogeneity from video data is difficult and hence requires further clustering and sampling techniques.
To facilitate the same, 80 videos were taken using "Sony Cyber-shot DSC-T77" 10.1 MP camera having resolution 640 x 480. Each video was of time duration less than 2 minutes approximately with 30 frames/second; out of them few had to be discarded manually as they did not contain the required objects. For data collection, a camera was mounted on a tripod which was periodically moved at different pan and tilt angles for posture variations. Data was collected at different spots and timings. However the amount of data from each scenario is not uniform. For example, during morning, the traffic movement (for example cars, pedestrians,two wheelers etc) is minimal. There are a number of issues that need to be considered. Some of these include blurring of picture due to the apparent motion of the tripod.
\begin{enumerate}
  \item Data Preprocessing: The aim of pre-processing is an enhancement of the image data that suppresses undesired distortions or enhances some relevant features of image for further processing and analysis task. General preprocessing techniques utilized by researchers include digital spatial filtering, intensity distribution linearization, contrast enhancement etc \citep{bernal2013impact}. In general, these methods are particularly essential for preprocessing while dealing with foreground extraction. For CNN based feature extraction, the data is normalized by zero centering and/or dividing by standard deviation before it is fed to the network. In our case, we use two different pre-processing techniques. First, when we are selecting data for clustering(for non-homogeneous data set) and second when it is being fed to the neural network. For the first case, we use FFT based specification technique to stabilize the color component. For this one frame is selected as base frame and intensity of rest of the frames is equalized according to the base using FFT as shown in equation \ref{fftt}. Later is done according to the traditional way of preprocessing.
      \begin{eqnarray}\label{fftt}
        X(k)=\sum_{n=0}^{N-1}x(n)W_{N}^{kn},    0\leq k \leq N-1 \\
        W_{N}=e^{-j\frac{2\pi}{N}} \nonumber
      \end{eqnarray}
  \item Data Set Sampling: It is a known fact that a good training set is one which represents diverse information. Hence data homogeneity resulting from video data (specially high frame rate) may impact the performance. We deal with the issue via a 2 step procedure. 1) Key-frame selection\citep{gao2017key} is employed to select candidate key frames. The number of clusters is same as the number of key frames in a video. 2) K-means clustering\citep{chao2018discriminative} is used on deep CNN features to get close clusters, cluster centre of each cluster is selected. VGG net trained on ILSVRC-2012 is used for extracting features in both cases. These algorithms group a set of objects in such a way that objects in the same cluster are more similar to each other than to those in other groups. Finally random clusters were taken as samples of data. Labelling of data is done with the help of human annotation. 16 object classes considered are listed as \{Aeroplane,Bus,Bicycle,Boat,Car,Cat,Dog,Horse,Motorbike,Person,Plotted plant,Sheep,Train and Background\}.
\end{enumerate}
%
%
%

\section{Experiments}
While performing object detection experiments, it was observed that object detection using RCNN, fast RCNN and faster RCNN did not give effective results on our own collected dataset. However,YOLO provides better results compared to the former 3 as shown in Table \ref{tab1}. It should be noted that the amount of false alarms was alarming for RCNN, fast and faster as compared to YOLO as shown in Table \ref{tabbb}.
\begin{table}[]
\centering
\begin{tabular}{|l|c|c|c|}
\hline
\multirow{2}{*}{Method/testset} & \multicolumn{3}{l|}{Accuracy} \\ \cline{2-4}
                                & CTS      & OTS     & PTS      \\ \hline
RCNN                            & 40       & 54      & 60.8     \\
Fast RCNN                       & 54       & 61      & 68.7     \\
Faster RCNN                     & 60       & 66      & 69.9     \\
Yolo                            & 65       & 68      & 74       \\ \hline
\end{tabular}
\caption{Results of different test sets from different pre-trained networks} \label{tab1}
\end{table}
\begin{table}
\centering
\begin{tabular}{|l|c|c|c|}
  \hline
  Method/testset & OTS \\
  \hline
  RCNN & 35 \\
  Fast RCNN & 29  \\
  Faster RCNN & 23  \\
  Yolo & 15 \\
  \hline
\end{tabular}
\caption {Results of false alarms of own test set from different pre-trained networks} \label{tabbb}
\end{table}
As seen from the Table \ref{tab1}, we have utilized three datasets referred as CTS, OTS and PTS. CTS is the subset of Caltech dataset. We refer our own test set as OTS while PTS is the subset of PASCAL VOC 2007 dataset. Results of table \ref{tab1} and 2 are shown using 1200 images of each set.

We further performed the following experiments(A to D) using CTS, OTS and PTS. It should be noted that for the different experiments (discussed below), training was done using 1300000 images of PTS. Around 300 region proposals were extracted from each image of every set using RPN. Test results shown in Tables \ref{tab2} to \ref{tablast} were conducted using CTS, OTS and PTS with 1200 images of each set. To be at par with results presented in different papers, we used SVM classification of features extracted from the second last layer of the trained nets. \\
A.	Layer wise: Features extracted from VGG(last max pool('pool5',$32^{nd}$) layer from 41 layered network), for
    dataset PASCAL VOC was to used train NoCs with different architectures.\\
B.	Learning rate wise: We use different learning rate patterns on the data to compare the effect.\\
C.	Dataset wise: Experiments were done on normal as well as blur datasets. These were then tested with blurred as
    well as general networks.\\
D.	Feature wise: Optical features were extracted from the images and orientation function was used to enhance the
    features of images and these were used for training the different networks with different data.\\

\begin{figure}
   \begin{center}
       \includegraphics[width=\linewidth]{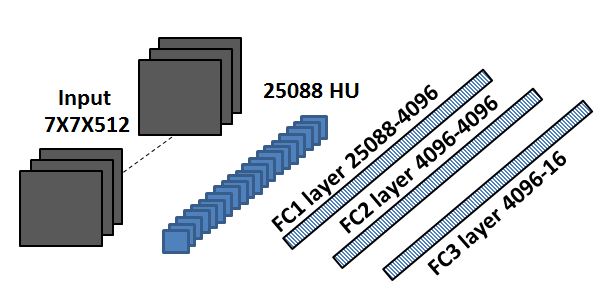}
    \caption{Architecture of $0C3fc$(FC-Fully Connected Layer,HU-Hidden Units)}\label{arch3}
    \end{center}
\end{figure}

\begin{figure}
   \begin{center}
       \includegraphics[width=\linewidth]{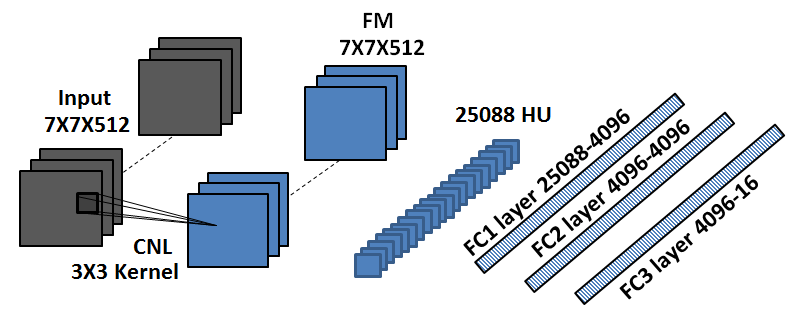}
    \caption{Architecture of $1C3fc$(CNL-Convolutional layer used with RELU, FC-Fully Connected Layer,FM-Feature Map}\label{arch2}
    \end{center}
\end{figure}

\begin{figure}
   \begin{center}
       \includegraphics[width=\linewidth]{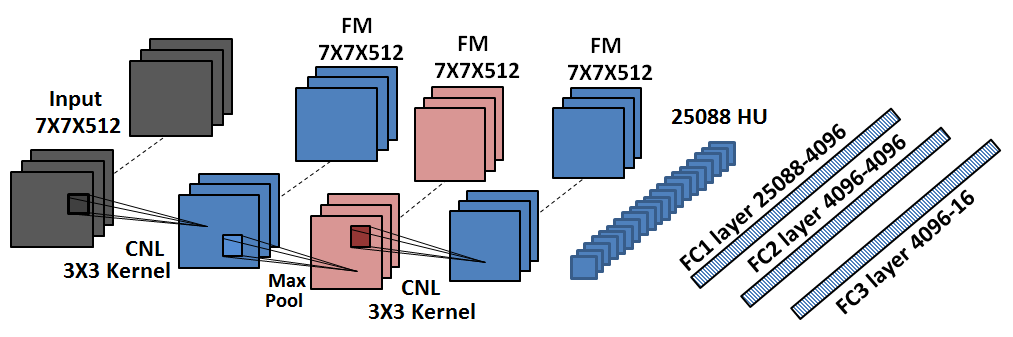}
    \caption{Architecture of $1M1$(CNL-Convolutional layer used with RELU, FC-Fully Connected Layer,FM-Feature Map)}\label{arch1}
    \end{center}
\end{figure}

A.	Three different architectures were used to develop the CNN based classifier. Figure \ref{arch3} depicts the widely used multiple fc layers for classification (referred as $0C3fc$). The second architecture as shown in Figure \ref{arch2} used 1 spatial convolutional followed by 3 fc layers (referred as $1C3fc$). The third architecture used 2 convolutional layers as represented in Figure \ref{arch1}. This is henceforth called $1M1$. Experiments performed using these NoCs with different test sets are shown in Table \ref{tab2}.
\begin{table}
\centering
\begin{tabular}{|l|c|c|c|}
\hline
\multirow{2}{*}{Method}                   & \multicolumn{3}{|c|}{Accuracy} \\ \cline{2-4}
                                          & CTS     & OTS      & PTS      \\ \hline
f4096-f4096-f16                           & 70      & 65       & 66       \\
1conv-f4096-f4096-f16(1C3fc)              & 83      & 81       & 75       \\
1conv-1maxPool-1conv-f4096-f4096-f16(1M1) & 78      & 76.6     & 73.8     \\ \hline
\end{tabular}
\caption{Accuracy of different test sets with different NoCs}
\label{tab2}
\end{table}

B.	Learning rate plays an important role for the convergence of training loss. RMSProp uses Hessian-based pre-conditioning with first order gradients for adaptive learning rates. However, it is important to effectively handle noise included in first order gradients during stochastic optimization (mini batch settings). Other variants of RMSProp such as AdaDelta and Adane is also considered superior to SGD in terms of training speed based on the fact that saddle points will slower the progress of first order gradients. SGD iteratively updates the parameter $\theta$ as shown in equation \ref{normal}.
\begin{equation}\label{normal}
  \theta_{t}=\theta_{t-1}-\alpha\nabla f(\theta_{t-1})   where
\end{equation}
$\alpha$ is the learning rate and $\nabla f(\theta_{t-1})$ is the first order gradient. The updating value of RMSProp is given in equation \ref{normal1}.
\begin{eqnarray}\label{normal1}
  \theta_{t}=\theta_{t-1}-\frac{\alpha}{\sqrt{\psi_{t}}+\varepsilon}\nabla f(\theta_{t-1})   where\\
  \psi_{t}=\beta\psi_{t-1}+(1-\beta)\nabla f(\theta_{t-1})^{2} \nonumber
\end{eqnarray}
In Hessian based conditioning, the training efficiency is increased by reducing the hessian condition number by transforming the parameters as represented in equation \ref{ab}.
\begin{equation}\label{ab}
  \theta_{t}=\theta_{t-1}-\alpha D^{-1} \nabla f(\theta_{t-1})
\end{equation}
Here $D=\sqrt{diag(H^{2})}$, which works even when H is indefinite as is the case for saddle points. It is verified in \citep{dauphin2015equilibrated} that $\sqrt{\psi_{t}}$ can be used as $D$.\\
\citep{ida2017adaptive} used a covariance matrix based pre-conditioning to deal with noisy gradients in mini-batches. They argued that if covariance $c[i,j]$ has a large value then the gradient strongly oscillates leading to inefficient progress of updating directions. The gradients are pre-conditioned as shown in equation \ref{cd} , covariance $c_{t}^{2}$ and $\mu_{t}$ mean are given in equation \ref{ef}.
\begin{equation}\label{cd}
  \theta_{t}=\theta_{t-1}-\frac{\alpha}{\sqrt{c_{t}^{2}}+\varepsilon}\nabla f(\theta_{t-1})   where
\end{equation}
\begin{eqnarray}\label{ef}
  c_{t}^{2}=\gamma c_{t-1}^{2}+\gamma(1-\gamma)(\nabla f(\theta_{t-1})-\mu_{t-1})^{2} \nonumber\\
  \mu_{t}=\gamma\mu_{t-1}+(1-\gamma)\nabla f(\theta_{t-1})
\end{eqnarray}
 We have divided PASCAL VOC 2007 dataset into three parts (DS1, DS2, DS3). Different methods were used to train the networks like:
\begin{enumerate}
  \item We train DS1 for 100 iterations. The trained net is then used to train DS2 which is further used for DS3. In all the cases, learning rate of linear decay from 0.01 to 0.005 was used and weights were updated according to equation \ref{normal}. This is referred as 1LR.
  \item In this case, we use net trained with DS1 for DS2 in the same way, but weight updation was done according to equation \ref{normal1}. We name it as 2LR.
  \item In this case, we train DS1,DS2 and DS3 (referred as j=1,2,3) and update weights as discussed in equation \ref{ab} rewritten as equation \ref{gh}. Final weights $\theta_{t}^{f}$ were updated as shown in equation \ref{ij}. This process is abbreviated as 3LR. The results for the same are depicted in Table \ref{tab3}. As observed from these results, $1C3fc$ NoC gave better results with 3LR.
      \begin{equation}\label{gh}
         \theta_{t}^{j}=\theta_{t-1}^{j}-\frac{\alpha}{\sqrt{c_{t}^{2}}+\varepsilon} \nabla f(\theta_{t-1}^{j})
      \end{equation}

      \begin{equation}\label{ij}
        \theta_{t}^{f}=\frac{1}{3}\sum_{j=1}^{3}(\theta_{t})
      \end{equation}
\end{enumerate}
\begin{table}
\centering
\begin{tabular}{|l|c|c|c|c|}
\hline
\multirow{2}{*}{Method}                       & \multirow{2}{*}{Learning Rate} & \multicolumn{3}{c|}{Accuracy} \\ \cline{3-5}
                                              &                                & CTS      & OTS      & PTS     \\ \hline
\multirow{3}{*}{1conv-f4096-f4096-f16(1C3fc)} & 1LR                            & 79.4     & 80       & 73      \\
                                              & 2LR                            & 80       & 80       & 73      \\
                                              & 3LR                            & 83       & 81       & 75      \\ \hline
1conv-1maxPool-1conv-f4096-f4096-f16(1M1)     & 3LR                            & 78       & 76.6     & 73.8    \\ \hline
\end{tabular}
\caption{Accuracy of different test sets with different NoCs trained with different learning rates}
\label{tab3}
\end{table}

\begin{figure}
   \begin{center}
       \includegraphics[width=\linewidth]{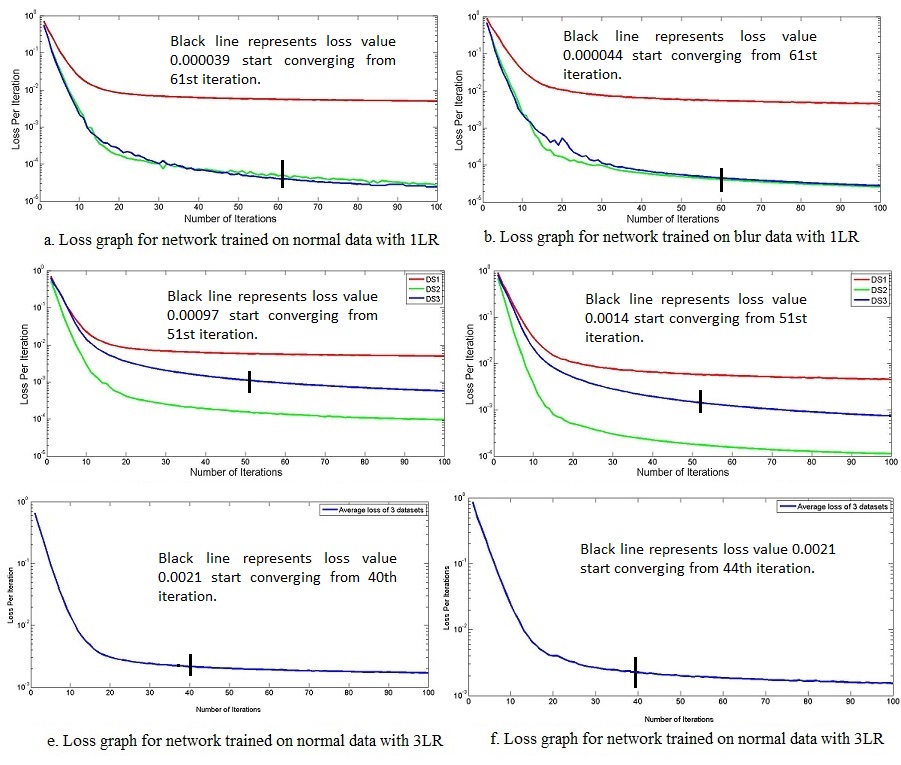}
    \caption{Losses for different learning methods for NoC ($1C3fc$) trained with normal and blur data.}\label{figloss1}
    \end{center}
\end{figure}

\begin{figure}
   \begin{center}
       \includegraphics[width=\linewidth]{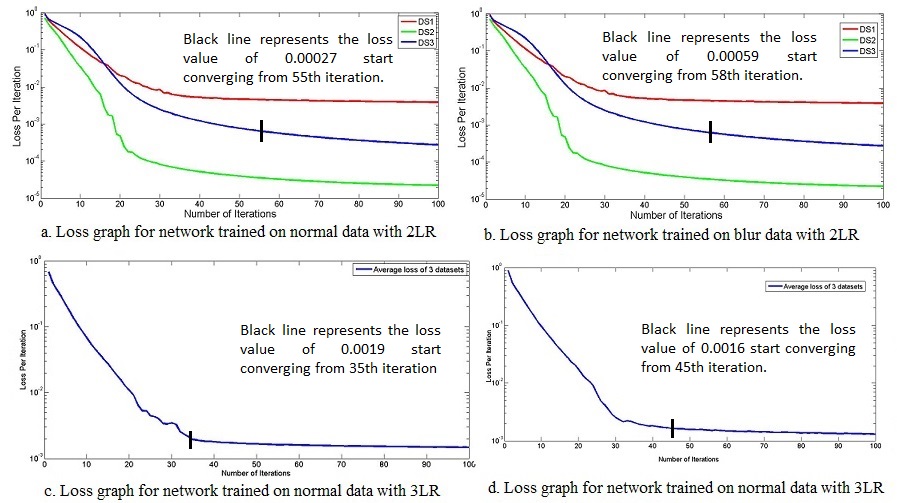}
    \caption{Losses for learning methods(2LR and 3LR) for NoC ($1M1$) trained with normal and blur data.}\label{figloss2}
    \end{center}
\end{figure}

C.	The same experiments(A and B) were performed on blur data. When blurred data was given to the networks trained on normal images, they gave poor results of classification accuracies as shown in Table \ref{tab6}. Hence, the networks were trained using blurred images. Different combinations of results(accuracy) are presented in Tables \ref{tab4}, \ref{tab5} and \ref{tab7}. All these results are shown using $1C3fc$. These include features of unblurred(referred to as Normal)/blur data extracted from last layer of net trained with Normal/blur data. These extracted features were used for testing purpose by giving them as input to SVM trained on normal/blur data. The various combinations are listed below:\\
1.	Normal data, normal net and SVM trained on normal data (N-N-N).\\
2.	Normal data, blur net and SVM trained on blur data (N-B-B).\\
3.  Blur data, normal net and SVM trained on normal data (B-N-N).\\
4.	Blur data, blur net and SVM trained on blur data (B-B-B).\\
From all the above results, it was seen that blur data does not give good results when tested using net trained on normal data. However normal data performs similarly on both normal as well as blur net.
 The losses obtained from networks having 1 and 2 convolutional layers trained with normal and blur data with different learning rates are represented in Figure \ref{figloss1} and \ref{figloss2} respectively. It is seen that the training loss converges better for 1LR and 3LR as compared to 2LR for both $1C3fc$ and $1M1$. The training losses and t-SNE plots along with the test accuracies also point towards the inference that 3LR with $1C3fc$ is the most suitable among the different options considered here.
\begin{table}
\centering
\begin{tabular}{|l|c|c|c|}
\hline
Learning Rate/testsets & CTS  & OTS & PTS \\ \hline
1LR                    & 59.4 & 68  & 60  \\
2LR                    & 58.2 & 68.5  & 56  \\
3LR                    & 59   & 69  & 62  \\ \hline
\end{tabular}
\caption{B-N-N}
\label{tab6}
\end{table}
\begin{table}
\centering
\begin{tabular}{|l|c|c|c|}
\hline
Learning Rate/testsets & CTS  & OTS & PTS \\ \hline
1LR                    & 79.4 & 80  & 73  \\
2LR                    & 80   & 80  & 73  \\
3LR                    & 83   & 81  & 74  \\ \hline
\end{tabular}
\caption{N-N-N}
\label{tab4}
\end{table}
\begin{table}
\centering
\begin{tabular}{|l|c|c|c|}
\hline
Learning Rate/testsets & CTS  & OTS  & PTS  \\ \hline
1LR                    & 70   & 73   & 68.8 \\
2LR                    & 72.5 & 74.3 & 71.4 \\
3LR                    & 73   & 70   & 71.6 \\ \hline
\end{tabular}
\caption{N-B-B}
\label{tab5}
\end{table}
\begin{table}
\centering
\begin{tabular}{|l|c|c|c|}
\hline
Learning Rate/testsets & CTS  & OTS  & PTS  \\ \hline
2LR                    & 70.6 & 73   & 72.5 \\
3LR                    & 73.7 & 74.6 & 72   \\ \hline
\end{tabular}
\caption{B-B-B}
\label{tab7}
\end{table}
\begin{figure}
   \begin{center}
       \includegraphics[width=\linewidth]{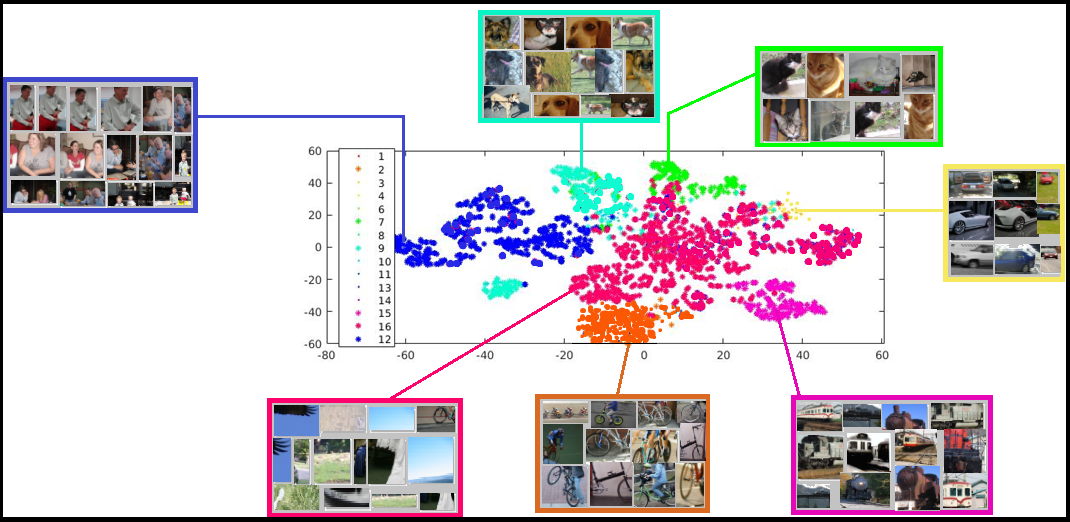}
    \caption{t-SNE distribution for the subset of training data extracted from $1C3fc$ with (3LR)}\label{tsne}
    \end{center}
\end{figure}
\begin{figure}
   \begin{center}
       \includegraphics[width=\linewidth]{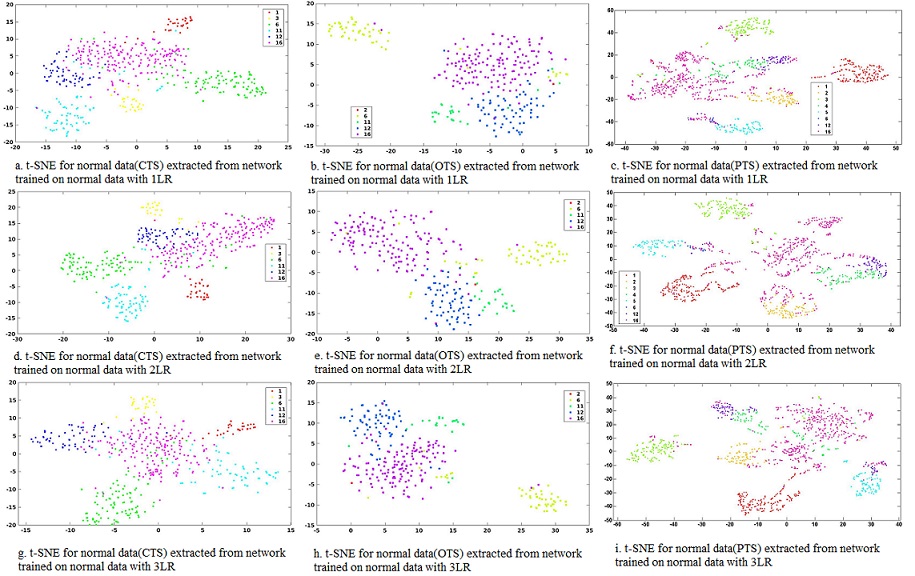}
    \caption{t-SNE for normal data extracted from NoC ($1C3fc$) trained on normal data with different learning methods}\label{tsne1}
    \end{center}
\end{figure}
\begin{figure}
   \begin{center}
       \includegraphics[width=\linewidth]{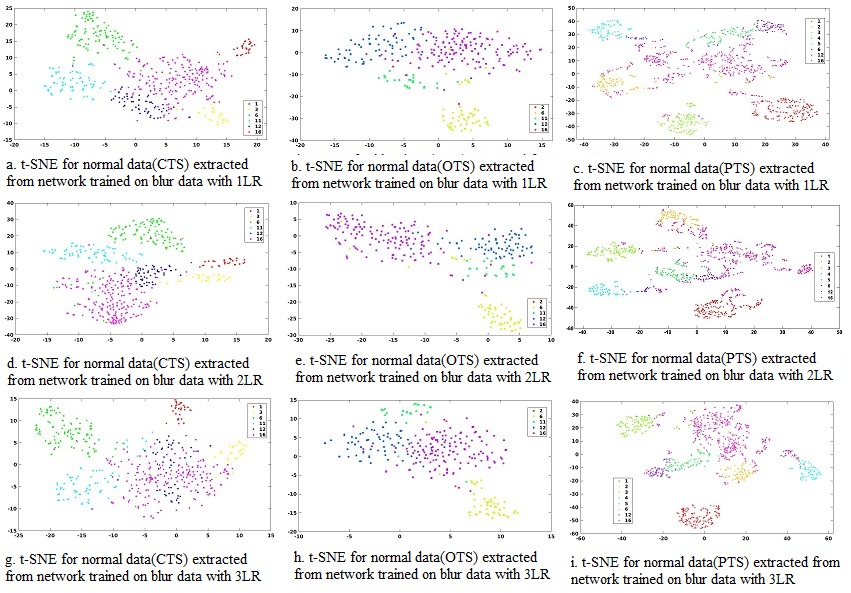}
    \caption{t-SNE for normal data extracted from NoC ($1C3fc$) trained on blur data with different learning methods}\label{tsne2}
    \end{center}
\end{figure}
\begin{figure}
   \begin{center}
       \includegraphics[width=\linewidth]{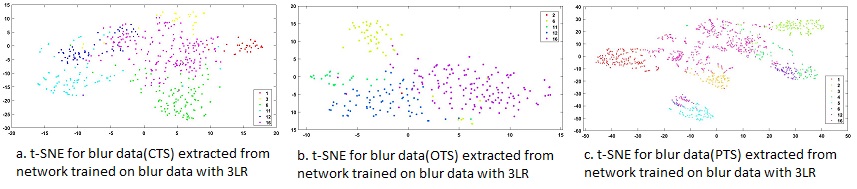}
    \caption{t-SNE for blur data extracted from NoC ($1C3fc$) trained on blur nets 3LR learning methods}\label{tsne3}
    \end{center}
\end{figure}
\begin{figure}
   \begin{center}
       \includegraphics[width=\linewidth]{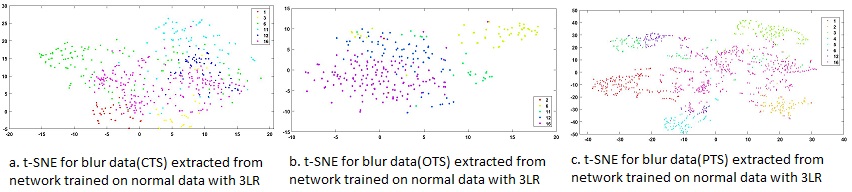}
    \caption{t-SNE for blur data extracted from NoC ($1C3fc$) trained on normal nets 3LR learning methods}\label{tsne4}
    \end{center}
\end{figure}
\begin{figure}
   \begin{center}
       \includegraphics[width=\linewidth]{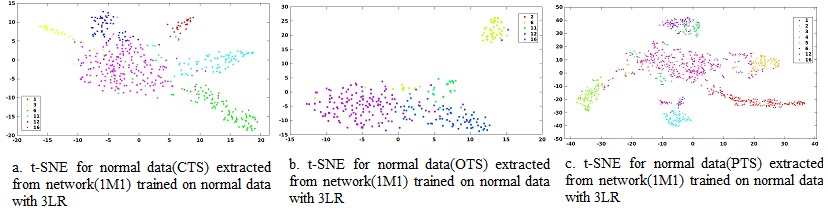}
    \caption{t-SNE for normal data extracted from NoC ($1M1$) trained on normal nets 3LR learning methods}\label{tsne5}
    \end{center}
\end{figure}
Test data are represented using t-SNE that is t-distributed Stochastic Neighbor Embedding which is defined as an algorithm for dimensionality reduction and is adapted to visualizing high-dimensional data in a scatter plot. The idea is to embed high-dimensional points into 2 or 3 dimensions in a manner that similarities among points retain. Nearby points in the high-dimensional space correspond to nearby embedded low-dimensional points, and distant points in high-dimensional space correspond to distant embedded low-dimensional points. To show the data distribution graphically, t-SNE for subset of training data is presented in Figure \ref{tsne} and for NoC with different datasets are depicted in  Figures \ref{tsne1}, \ref{tsne2}, \ref{tsne3}, \ref{tsne4} and \ref{tsne5}.
t-SNE for OTS in case of $1C3fc$ with 3LR gives good and separated clusters for every class.\\

\begin{figure}
  \centering
  \includegraphics[width=\linewidth]{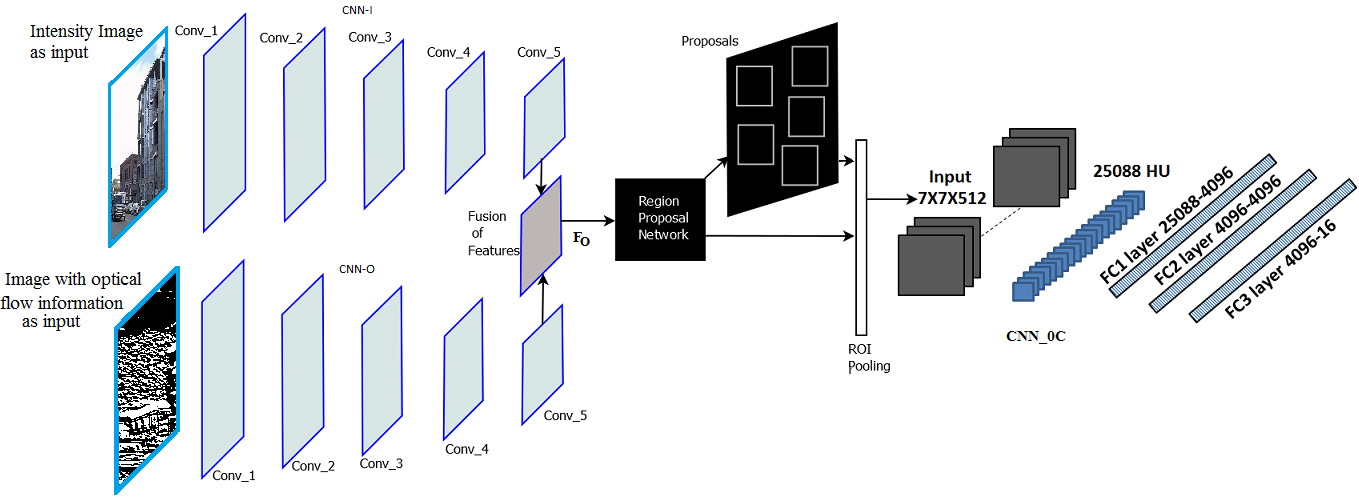}\\
  \caption{Multimodal object detection and classification using RGB and optical flow features}\label{O}
\end{figure}

D.	In this paper, features of intensity images extracted from 5th convolutional layer $Conv5\_F_{I}$ are fused (added) to orientation features $(conv5\_F_{O})$ extracted using optical flow. Feature map $F_{O}$ is obtained as shown in equation \ref{eqO}. The whole process is shown in Figure \ref{O} with classifier network $1C3fc$. Accuracy of only RGB and RGB+OF are shown in \ref{tabNN}.
      \begin{equation}\label{eqO}
        F_{O}=conv5\_F_{I}+conv5\_F_{o}
      \end{equation}
 Same experiments were performed for blur images as presented in Tables \ref{tabBN}, \ref{tabNB} and \ref{tabBB}.
\begin{table}
\centering
\begin{tabular}{|c|c|c|c|c|c|c|c|}
\hline
\multirow{3}{*}{Method} & \multirow{3}{*}{Learning Rate/testsets} & \multicolumn{6}{l|}{TEST SETS}                                                 \\ \cline{3-8}
                        & \multirow{2}{*}{}      & \multicolumn{2}{l|}{CTS} & \multicolumn{2}{l|}{OTS} & \multicolumn{2}{l|}{PTS} \\ \cline{3-8}
                        &                        & OF         & RGB         & OF          & RGB        & OF          & RGB        \\ \hline
1conv-f4096-f4096-f16   & 3LR                    & 84         & 83          & 79.8        & 81         & 74.8        & 74         \\ \hline
\end{tabular}
\caption{Accuracy of NoC with optical features (Normal dataset $\rightarrow$ normal nets $\rightarrow$ normal SVM)}
\label{tabNN}
\end{table}
\begin{table}
\centering
\begin{tabular}{|c|c|c|c|c|c|c|c|}
\hline
\multirow{3}{*}{Method} & \multirow{3}{*}{Learning Rate/testsets} & \multicolumn{6}{l|}{TEST SETS}                                                 \\ \cline{3-8}
                        & \multirow{2}{*}{}      & \multicolumn{2}{l|}{CTS} & \multicolumn{2}{l|}{OTS} & \multicolumn{2}{l|}{PTS} \\ \cline{3-8}
                        &                        & OF         & RGB         & OF          & RGB        & OF          & RGB        \\ \hline
1conv-f4096-f4096-f16   & 3LR                    & 58.4         & 59          & 64.6        & 69         & 60.5        & 62\\ \hline
\end{tabular}
\caption{Accuracy of NoC with optical features (Blur dataset $\rightarrow$ normal nets $\rightarrow$ normal SVM)}
\label{tabBN}
\end{table}
\begin{table}
\centering
\begin{tabular}{|c|c|c|c|c|c|c|c|}
\hline
\multirow{3}{*}{Method} & \multirow{3}{*}{Learning Rate/testsets} & \multicolumn{6}{l|}{TEST SETS}                                                 \\ \cline{3-8}
                        & \multirow{2}{*}{}      & \multicolumn{2}{l|}{CTS} & \multicolumn{2}{l|}{OTS} & \multicolumn{2}{l|}{PTS} \\ \cline{3-8}
                        &                        & OF         & RGB         & OF          & RGB        & OF          & RGB        \\ \hline
1conv-f4096-f4096-f16   & 3LR                    & 75.8         & 73          & 73.3        & 70         & 72        & 71.6\\ \hline
\end{tabular}
\caption{Accuracy of NoC with optical features (Normal dataset $\rightarrow$ blur nets $\rightarrow$ blur SVM)}
\label{tabNB}
\end{table}
\begin{table}
\centering
\begin{tabular}{|c|c|c|c|c|c|c|c|}
\hline
\multirow{3}{*}{Method} & \multirow{3}{*}{Learning Rate/testsets} & \multicolumn{6}{l|}{TEST SETS}                                                 \\ \cline{3-8}
                        & \multirow{2}{*}{}      & \multicolumn{2}{l|}{CTS} & \multicolumn{2}{l|}{OTS} & \multicolumn{2}{l|}{PTS} \\ \cline{3-8}
                        &                        & OF         & RGB         & OF          & RGB        & OF          & RGB        \\ \hline
1conv-f4096-f4096-f16   & 3LR                    & 77         & 73.7          & 75        & 74.6         & 72.5        & 72\\ \hline
\end{tabular}
\caption{Accuracy of NoC with optical features (Blur dataset $\rightarrow$ blur nets $\rightarrow$ blur SVM)}
\label{tabBB}
\end{table}

\begin{figure}
   \begin{center}
       \includegraphics[width=\linewidth]{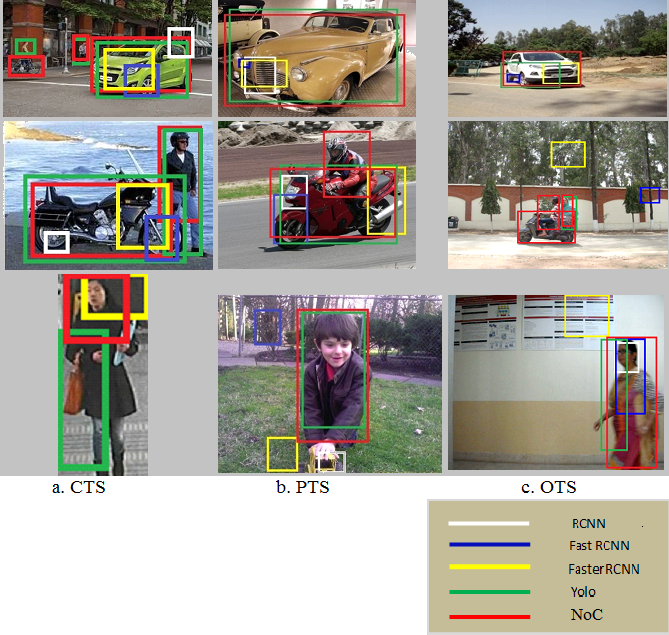}
    \caption{Comparison of our NoC with other object detection method}\label{res}
    \end{center}
\end{figure}

\clearpage
 Figure \ref{res} shows comparison of detection between RCNN, fast RCNN, faster RCNN, yolo and NoC($1C3fc$). Table \ref{tablast} shows accuracy of detection with all these methods.
\begin{table}
\centering

\begin{tabular}{|l|c|c|c|}
\hline
Method/Test Sets & CTS & OTS & PTS  \\ \hline
RCNN             & 40  & 54  & 60.8 \\
Fast RCNN        & 54  & 61  & 68.7 \\
Faster RCNN      & 60  & 66  & 69.9 \\
Yolo             & 65  & 68  & 74   \\
\textbf{NoC }             & \textbf{83}  & \textbf{81}  & \textbf{74}   \\ \hline
\end{tabular}
\caption{Comparison of accuracies of NoC with other object detection methods}
\label{tablast}
\end{table}
\section{Conclusion}
This paper discussed simple data collection and sampling tricks prior training. Extensive experiments are performed on different convolutional classification architecture, with various learning rates. Results depict that $1C3fc$ with 3LR gives relatively better performance. We also use blur data to train these NoCs. It is observed that blur net can be used for blurred as well as unblurred data whereas network trained with normal data fails to tackle blurred data. Further optical flow features computed for training normal as well as blurred NoCs prove to be beneficial. Pre-conditioning with first order gradients for adaptive learning rates is also utilized to deal with saddle points. $3LR$ outperforms the others in terms of training loss convergence with early iteration.

\bibliography{3rdpaperref}



\par\noindent
\parbox[t]{\linewidth}{
\noindent\parpic{\includegraphics[height=1.5in,width=1in,clip,keepaspectratio]{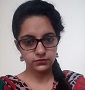}}
\noindent {\bf Baljit Kaur}\
is pursuing her Ph.D. in Computer Science department at Thapar Institute of Engg and Tech, Patiala. She received her M.Tech degree in Information Technology from Guru Nanak Dev University, Amritsar and B.Tech degree in Computer Science from Amritsar College of Engg and Technology, Amritsar. She has five years of teaching experience. Her research area is image processing focused on augmented map based intelligent navigation system.}
\vspace{2\baselineskip}
\par\noindent
\parbox[t]{\linewidth}{
\noindent\parpic{\includegraphics[height=1.5in,width=1in,clip,keepaspectratio]{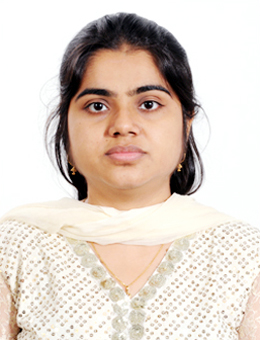}}
\noindent {\bf Jhilik Bhattacharya}\
works as an assistant professor in Computer Science department at Thapar Institute of Engg and Tech, Patiala. She received her Ph.D. degree in Computer Science from NIT, Durgapur. She has 10 years of research and teaching experience.Her research interests include image processing,computer vision, pattern recognition.}

\end{document}